\documentclass[10pt, a4paper]{article}

\usepackage{amsmath, verbatim, amssymb, booktabs}
\usepackage{stfloats}
\usepackage{hyperref}

\usepackage[final]{lrec2026} 

\title{``OK Aura, Be Fair With Me'': Demographics-Agnostic Training for Bias Mitigation in Wake-up Word Detection}

\name{\begin{tabular}[t]{c}
Fernando López$^{1,2}$, Paula Delgado-Santos$^{1}$ \\
Pablo Gómez$^{1}$, David Solans$^{1}$, Jordi Luque$^{1}$
\end{tabular}}

\address{$^{1}$Telefónica Innovación Digital, Madrid, Spain \\
         $^{2}$Universidad Autónoma de Madrid, Madrid, Spain \\
         \{fernando.lopez,  paula.delgadodesantos, pablo.gomezguerrero,\\ david.solansnoguero, jordi.luque\}@telefonica.com\\}
         
\abstract{
Voice-based interfaces are widely used; however, achieving fair Wake-up Word detection across diverse speaker populations remains a critical challenge due to persistent demographic biases. This study evaluates the effectiveness of demographics-agnostic training techniques in mitigating performance disparities among speakers of varying sex, age, and accent. We utilize the OK Aura database for our experiments, employing a training methodology that excludes demographic labels, which are reserved for evaluation purposes. We explore (i)  data augmentation techniques to enhance model generalization and (ii) knowledge distillation of pre-trained foundational speech models. The experimental results indicate that these demographics-agnostic training techniques markedly reduce demographic bias, leading to a more equitable performance profile across different speaker groups. Specifically, one of the evaluated techniques achieves a Predictive Disparity reduction of 39.94\% for sex, 83.65\% for age, and 40.48\% for accent when compared to the baseline. This study highlights the effectiveness of label-agnostic methodologies in fostering fairness in Wake-up Word detection.
 \\ \newline  }

\begin{document}

\maketitleabstract

\section{Introduction}
\label{sec:introduction}

Voice-based interfaces are now central to human-computer interaction, enabling virtual assistants, hands-free messaging, and applications such as customer support and clinical/legal transcription. The entry point to most of these systems is a Wake-up Word (WuW); a predefined trigger phrase that, once detected by an always-on lightweight acoustic model, activates the device and initiates interaction with the user~\citep{10.1117/12.666025, Kepuska11, lopez2023robust, lopez2021deep}. However, speech systems often exhibit performance disparities across demographic groups such as sex, age, and accent~\citep{attanasio2024twists}. Acoustic variability can systematically affect and raise fairness concerns~\citep{fuckner2023uncovering}. In always-on settings, these disparities manifest not only as aggregate error-rate gaps but as unequal \emph{interactional burdens}~\citep{Choi_Choi_2025}: some users must repeat commands or alter their speech more often than others to obtain the same functionality. WuW detection is particularly susceptible because decisions rely on short speech segments with limited context, which can amplify speaker-dependent variability and reduce reliability for children, older adults, and regional or non-native speakers.

These concerns are consistent with prior work documenting demographic bias across speech tasks, including speaker identification, phoneme recognition, intent classification, keyword spotting (KWS), and emotion recognition (ER)~\citep{meng2022dont, hutiri2023tiny, slaughter2023pre}. In Automatic Speech Recognition (ASR), higher Word Error Rates (WER) are repeatedly reported for speakers with regional or non-native accents, with additional disparities linked to sex, age, and intersectional factors~\citep{garg2018word, zolnoori2024decoding, harris2024modeling, feng2021quantifying, martin2023bias}. Evaluations of large foundation models such as Whisper~\citep{radford2023robust} further corroborate persistent racial, sex, and dialect biases, frequently favoring majority or privileged groups~\citep{fuckner2023uncovering, slaughter2023pre, hutiri2023tiny}. Similar patterns have been observed in KWS and ER, where systems often underperform for children, elderly speakers, and nonstandard accents~\citep{mujtaba2024lost, hutiri2023tiny, feng2021quantifying, martin2023bias}. Recent benchmark efforts such as Fair-Speech for ASR~\citelanguageresource{veliche2024towards} and FaiST for broader speech technology~\citelanguageresource{jahan2025faist} further document systematic performance gaps across multiple demographic attributes, underscoring the need for dedicated fairness analyses in speech interfaces.

Several methodological tools have been proposed to diagnose and mitigate these disparities. For instance, DivExplorer can automatically identify attribute combinations (e.g., sex, age, accent) associated with large performance gaps~\citep{pastor2021looking}. Building on such diagnostics, mitigation frameworks such as CLUES use discovered subgroups to guide contrastive learning and reduce disparities by targeting underperforming cohorts in the representation space~\citep{koudounas2024contrastive}. In parallel, \citet{slaughter2023pre} demonstrated that embeddings from pre-trained speech models, including Whisper, wav2vec~2.0, WavLM, and HuBERT, can encode and amplify social biases. To measure this directly within the embedding spaces, they developed the Speech Embedding Association Test (SpEAT). Building on this area of research, \citet{lin2024sslbias} examined how specific architectural and data choices in self-supervised learning (SSL) impact social biases in downstream tasks.

Beyond specific methods and datasets, recent work argues that speech recognition fairness is inherently context-dependent and multi-metric rather than ``one-size-fits-all'', calling for benchmarks that consider task requirements, deployment constraints, and stakeholder needs~\citep{elghazaly2025fairasr, veliche2024towards}. 


Nonetheless, many mitigation strategies depend on explicit demographic labels, which are often unavailable, incomplete, or privacy-sensitive in real-world deployments, and data scarcity for underrepresented groups remains a persistent obstacle~\citep{dheram2022toward, barocas2016big}. Other approaches pursue fairness via personalization, for instance, by conditioning KWS models on speaker-specific embeddings to improve performance for underrepresented users~\citep{zhao2025personalizingkws}. While effective, such methods typically require additional user data and enrollment procedures, and may not be feasible for compact, always-on WuW detectors operating entirely on device.

Motivated by these limitations, we study demographic bias in WuW detection and develop mitigation strategies that do not require demographic labels during training. We (i) quantify demographic disparities across sex, age, and accent using group-wise analyses and fairness metrics such as Predictive Disparity (PD) and Disparate Impact (DI), and (ii) investigate demographics-agnostic training methods based on generalization-oriented data augmentation and knowledge distillation/transfer from large self-supervised foundation speech models. Experiments on a real-world Spanish WuW dataset (``OK Aura'') show that these label-free strategies substantially reduce demographic bias while preserving overall detection performance.

\section {Mitigation Methodology}
\label{sec:methodology}
We implement a mitigation pipeline that remains demographics-agnostic during training, reserving demographic labels strictly for post-hoc bias evaluation. First, we identify bias within the dataset to identify demographic groups underrepresented in the training and validation phases. Subsequently, we assess bias reflected in WuW classifier predictions. Then, we adopt demographics-agnostic training methodologies intended to alleviate those biases.


This choice is motivated by the high cost and practical barriers of collecting additional data for underrepresented groups (e.g., privacy and limited access). Even with balanced data, disparities can persist due to design choices~\citep{hutiri2023tiny} or feature selection~\citep{bailey2021gender}; while demographics-aware methods can reduce bias~\citep{dheram2022toward}, we target mitigation without explicit demographic conditioning.

Our methodology employs two demographics-unaware training strategies. First, we hypothesize that modulating or partially removing frequency information during training discourages the model from relying on demographic-correlated acoustic cues. Given that sex, age, and accent are known to correlate with F0 (fundamental frequency) and formant structure~\citep{vorperian2019corner}, spectral envelope~\citep{harnsberger2008speaking}, and prosody~\citep{piat2008foreign}, respectively. By disrupting these cues, the model could be encouraged to learn more invariant representations~\citep{vandenberghe2023augmentation}. To this end, we explore data augmentation techniques applied at the spectrum level (Section~\ref{sec:data-augmentation-techniques}). Second, large pre-trained SSL models, trained on diverse audio, have been shown to suppress speaker identity in their upper layers~\citep{mohamed2022self}. Furthermore, as some models have been scaled to encompass over 4 million hours of training data \citep{barrault2023seamless}, we hypothesize that they capture demographically robust representations. Therefore, we investigate using such models as teachers to train a compact, robust student model~\citep{chai2022fairness} (Section~\ref{sec:ssl_models}).

\subsection{Data augmentation techniques}
\label{sec:data-augmentation-techniques}


We consider both time-domain and time-frequency-domain augmentations. Given an input waveform $x \in \mathbb{R}^{N}$, we compute a time--frequency representation via the Short-Time Fourier Transform (STFT) and use its magnitude to form a spectrogram,
\begin{equation}
    X = \mathrm{Spectrogram}(x) = \left|\mathrm{STFT}(x)\right|^{2},
\end{equation}
where $X \in \mathbb{R}^{T \times F}$, $T$ denotes the number of time frames, and $F$ the number of frequency bins. For augmentations applied in the time-frequency domain, we modify the magnitude to obtain $X'$ while preserving the original phase, and then reconstruct an augmented waveform $x'$ using the inverse STFT (ISTFT).

\textbf{FreqMixStyle:} it mixes frequency-wise feature statistics between samples to promote domain-invariant representations. It is motivated by frequency-wise instance normalization analyses for audio domain generalization~\citep{kim2022domain}. In practice, we normalize a spectrogram $X_i$ along the frequency axis and re-scale it using mixed statistics from another randomly selected spectrogram $X_j$:
\begin{equation}
    \mu_{\text{new}} = \lambda \mu_i + (1 - \lambda)\mu_j,\quad
    \sigma_{\text{new}} = \lambda \sigma_i + (1 - \lambda)\sigma_j
\end{equation}
where $\lambda \sim \mathrm{Beta}(\alpha,\alpha)$ controls the interpolation strength.

\textbf{FilterAugment:} simulates acoustic filtering by applying smooth frequency-dependent gains rather than masking entire bands~\citep{nam2022filteraugment}. Given a spectrogram $X$, we apply a multiplicative weighting mask
\begin{equation}
    X' = X \odot W_{\mathrm{FA}},
\end{equation}
where $W_{\mathrm{FA}} \in \mathbb{R}^{T \times F}$ contains frequency-dependent weights and $\odot$ denotes element-wise multiplication. We use the linear variant, which linearly interpolates gains across frequency to avoid abrupt discontinuities~\citep{nam2022filteraugment}.

\textbf{Frequency masking:} as a strong baseline augmentation, we also apply frequency masking from SpecAugment~\citep{park2019specaugment}. We sample the mask width $f \sim \mathcal{U}(0,W_F)$ and starting index $f_0 \sim \mathcal{U}(0,\nu-f)$, where $\nu$ is the number of mel channels, and set the band $[f_0, f_0+f)$ to zero:
\begin{equation}
    X' = \mathrm{FreqMask}(X).
\end{equation}
This technique has been shown to improve model robustness by forcing the network to learn from partial spectrograms, thus improving generalization. It is especially effective in situations where the model must handle varying acoustic conditions or incomplete audio inputs \citep{kim2021specmix}.

\textbf{Device Impulse Responses (DIR).}
Impulse responses model how a capture device filters an input signal. Originally, it was presented for device generalization by simulating microphone characteristics. Nonetheless, it modulates frequencies by convolving each training utterance with a sampled device impulse response $h_{\mathrm{dir}}$~\citep{morocutti2023device}:
\begin{equation}
    x' = x \ast h_{\mathrm{dir}} .
\end{equation}
To keep input dimensions consistent, we truncate the convolved signal to match the original length.

\subsection{Speech Self-Supervised Learning models} \label{sec:ssl_models}


SSL has become a key approach for learning robust speech representations from large amounts of unlabeled audio~\citep{mohamed2022self, chen2022large, han2025leveraging, wang2024asvspoof}. In WuW settings, SSL models can be particularly useful under limited labeled data and challenging acoustic conditions~\citep{yu2023few, mork2024noise}.

\begin{figure}[h!]
    \centering
    \includegraphics[
        width=\columnwidth,
        trim=0.3cm 0.3cm 0.2cm 0.1cm,
        clip
    ]{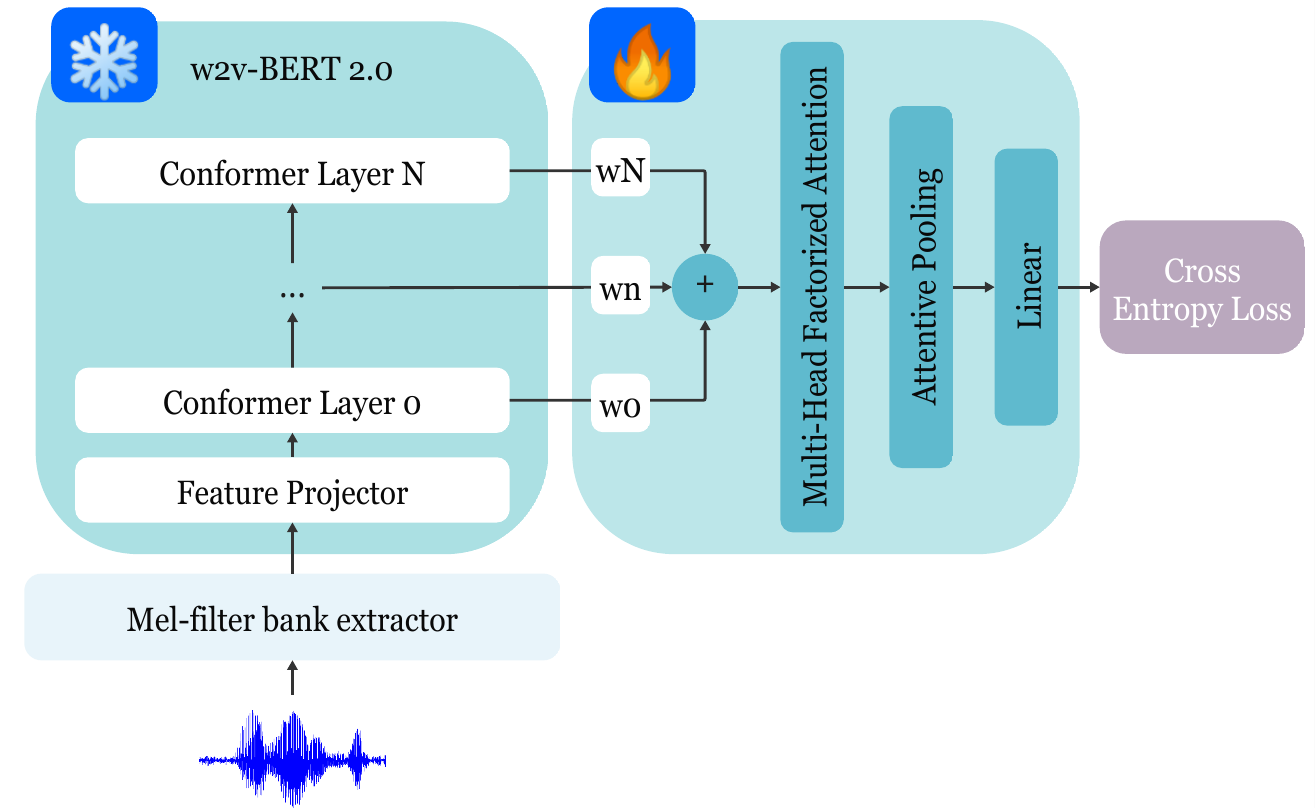}
    \caption{\texttt{w2v-BERT2-kws} architecture. Raw audio is converted to 80-channel Mel filterbanks, then passed through convolutional subsampling and a linear projection before a 24-layer Conformer encoder. Layerwise hidden states are combined via a learnable weighted sum, followed by Multi-Head Factorized Attention (MHFA), attentive pooling over time, and a linear classifier. The w2v-BERT~2.0 encoder is frozen; only the layer weights and the classification head are trained with cross-entropy.}
    \label{fig:ssl_classifier}
\end{figure}

We leverage a large SSL encoder to build a high-capacity teacher classifier (\texttt{w2v-BERT2-kws}) and then distill its knowledge into a compact WuW student model. Specifically, we use the w2v-BERT~2.0 pre-trained model~\citep{barrault2023seamless}, a Conformer-based multilingual speech encoder trained on 4.5M hours of unlabeled audio. The \texttt{w2v-BERT2-kws} architecture is depicted in Figure~\ref{fig:ssl_classifier}. We leverage the w2v-BERT~2.0 encoder frozen and train a lightweight classification head. Motivated by evidence that different transformer layers encode complementary information~\citep{pasad2023comparative}, we compute a learnable weighted sum over the 24 layerwise hidden states. The resulting sequence representation is then processed with Multi-Head Factorized Attention (MHFA), a parameter-efficient variant of multi-head attention that factorizes the attention projections so each head operates in a lower-dimensional subspace~\citep{peng2025mhfa}. Finally, the obtained result is summarized via attentive pooling before a final linear layer~\citep{peng2025mhfa,ronceldiaz24on}. The teacher is trained with cross-entropy and used exclusively for distillation; it is not intended for real-time, on-device inference.

\textbf{Knowledge distillation (KD):} after training the teacher, we freeze it and train the student via logit matching. The student minimizes a weighted combination of the standard cross-entropy (CE) loss with respect to ground-truth labels and a Kullback--Leibler (KL) divergence term between temperature-scaled teacher and student predictions:

\begin{equation}
\label{eq:distillation}
\begin{split}
L_{\text{KD}} =\;& \delta\, L_{\text{CE}}(p_{\text{student}}, y_{\text{true}}) \\
& + (1-\delta)\,\tau^{2}\,D_{\text{KL}}\!\bigl(p_{\text{teacher}}^{\tau}\parallel \log p_{\text{student}}^{\tau}\bigr)
\end{split}
\end{equation}

where $L_{\text{KD}}$ is the total loss function, and $\delta \in [0,1]$ is a weighting factor that controls the balance between the two loss components. $\tau$ is the temperature parameter that controls the sharpness of the probability distribution, and $L_{\text{CE}}(p_{\text{student}}, y_{\text{true}})$ is the cross-entropy loss, defined as:
    
\begin{equation}
    L_{\text{CE}}(p_{\text{student}}, y_{\text{true}}) = - \sum_{i} y_{\text{true}, i} \log p_{\text{student}, i}
\end{equation}

where the term $y_{\text{true}}$ represents the ground truth label and $p_{\text{student}}$ is the probability output from the student WuW model. Here $i$ refers to each specific class (WuW or unknown). The output probability is obtained by:

\begin{equation}
    p_{\text{student}, i} = \frac{e^{z_{\text{student}, i}}}{\sum_{j} e^{z_{\text{student}, j}}}
\end{equation}

where $z_{\text{student}, i}$ are the logits, and the $j$ index refers to the summation over both classes. To continue with, $D_{\text{KL}}(p_{\text{teacher}}^{\tau} \parallel p_{\text{student}}^{\tau})$ is the KL divergence between two temperature-scaled probability distributions:

\begin{equation}
    D_{\text{KL}}(p_{\text{teacher}}^{\tau} \parallel p_{\text{student}}^{\tau}) = \sum_{i} p_{\text{teacher}, i}^{\tau} \log \frac{p_{\text{teacher}, i}^{\tau}}{p_{\text{student}, i}^{\tau}}
\end{equation}

where $p_{\text{student}, i}^{\tau}$ is the probability output from the student WuW model, obtained by applying a temperature-scaled softmax:

\begin{equation}
    p_{\text{student}, i}^{\tau} = \frac{e^{z_{\text{student}, i} / \tau}}{\sum_{j} e^{z_{\text{student}, j} / \tau}}
\end{equation}

Similarly, $p_{\text{teacher}}$ is also temperature-scaled:

\begin{equation}
    p_{\text{teacher}, i}^{\tau} = \frac{e^{z_{\text{teacher}, i} / \tau}}{\sum_{j} e^{z_{\text{teacher}, j} / \tau}}
\end{equation}

where $z_{\text{teacher}, i}$ are the logits from the teacher model and $\tau$ is the temperature parameter. Higher $\tau$ produces softer targets, encouraging the student to match relative class confidences rather than only hard decisions.

\section{Datasets}
\label{sec:dataset}

We utilize a proprietary in-domain corpus, OK Aura (Section~\ref{sec:ok_aura}), and several publicly available out-of-domain resources for augmentation and robustness. Specifically, we incorporate Spanish Common Voice v7.1~\citelanguageresource{commonvoice_es_v7_1}, the M-AILabs Spanish corpus~\citelanguageresource{m_ailabs_speech_dataset}, real and simulated room impulse responses (RIRs) and noises from OpenSLR SLR28~\citelanguageresource{openslr_slr28_rirs_noises}, and environmental noise recordings from DEMAND~\citelanguageresource{demand}. To further improve device robustness, we additionally use microphone impulse-response collections including MicIRP\footnote{\url{https://micirp.blogspot.com/}} and the Multi-Angle Multi-Distance Microphone IR dataset~\citelanguageresource{multi_angle_multi_distance_micir_zenodo}.

\begin{figure}[h]
    \centering
    \includegraphics[width=1.0\linewidth]{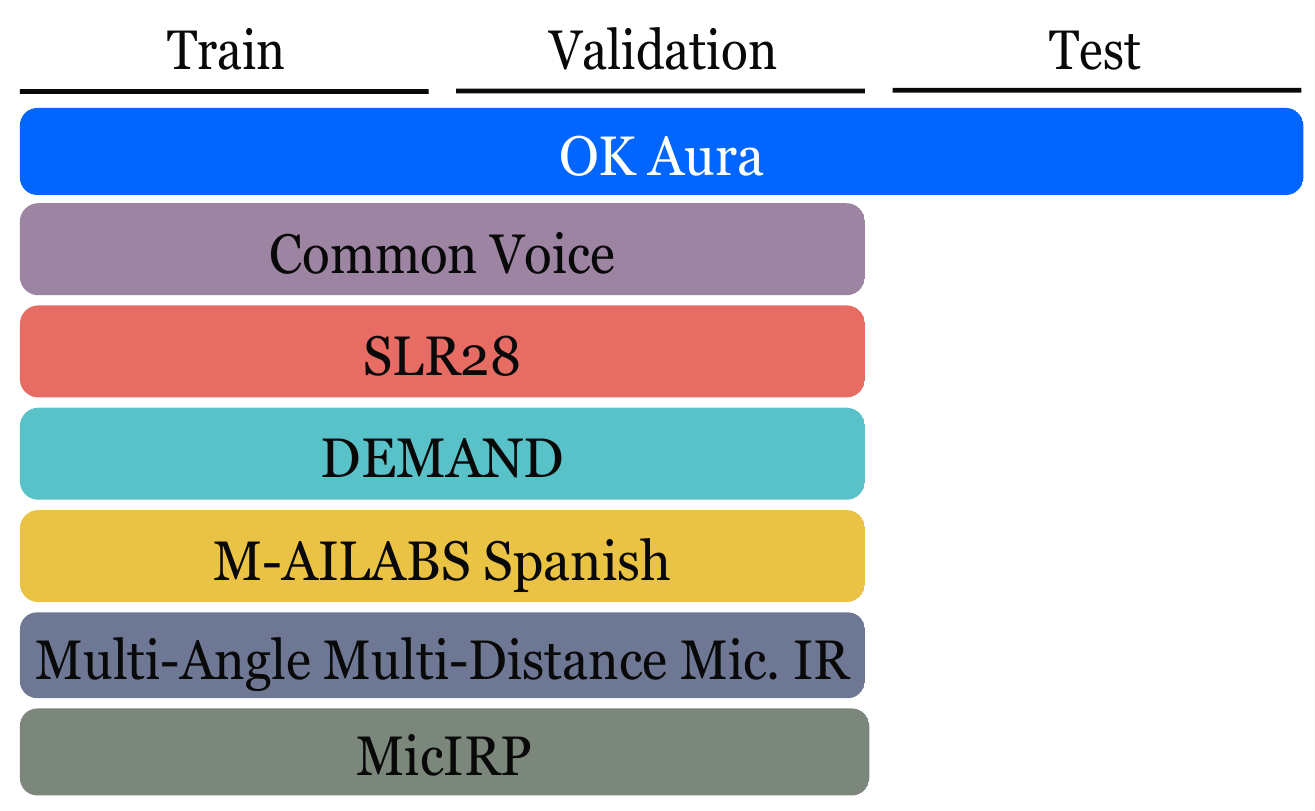}
    \caption{Dataset usage across train/validation/test splits. OK Aura is used in all phases (training, validation, and testing). Public resources are used primarily for training and validation, mainly for augmentation and robustness.}
    \label{fig:databases_usage}
\end{figure}

Figure~\ref{fig:databases_usage} summarizes how these resources are used across the experimental pipeline. OK Aura is used for training, validation, and testing. In contrast, the public corpora (Common Voice, M-AILabs, SLR28, DEMAND, MicIRP, and Multi-Angle Multi-Distance Microphone IR) are used primarily for training and validation to support augmentation and robustness. We restrict bias quantification and fairness evaluation to OK Aura because the out-of-domain datasets lack the necessary demographic metadata, provide insufficient granularity, or exhibit annotation mismatches relative to the WuW task.

\subsection{OK Aura Database}
\label{sec:ok_aura}

OK Aura contains approximately 5.8k audio samples ($\sim$4.5 hours) from 546 anonymized speakers. It includes both speech and non-speech material (e.g., background noise), and all speech content is in Spanish, comprising the wake-up phrase and additional utterances. The corpus provides speaker-level demographic annotations (sex, age, accent) and sample-level metadata including class labels (positive/negative), transcriptions for speech samples, and a sound-type tag (``speech'' vs.\ ``noise'').


To provide a concrete sense of the speech data, the corpus covers various realistic usage scenarios and challenging negative samples. Positive instances range from the isolated WuW (\textit{``OK Aura''}) to the WuW embedded within context sentences (e.g., \textit{``Perfecto, voy a mirar qué dan hoy. OK Aura''} `Perfect, I am going to see what is on today. OK Aura'). Additionally, negative samples include utterances with partial matches, such as containing only the word ``Aura'' (\textit{``Hay un aura de paz y tranquilidad.''} `There is an aura of peace and tranquility.') or ``OK'' (\textit{``OK, a ver qué ponen en la tele.''} `OK, let's see what is on TV.'). Furthermore, it includes distractors with words that sound similar to the target phrase, such as \textit{``Hola Laura,''} (`Hello Laura,'), \textit{``Prefiero el hockey al baloncesto,''} (`I prefer hockey to basketball,'), or a combination of both: \textit{``Porque Laura, ¿qué te pareció la película?''} (`Because Laura, what did you think of the movie?').

Furthermore, recordings span a wide range of acoustic environments as they were recorded in different spaces, from quiet rooms to natural background noise scenarios, and recorded across different devices. The dataset also includes temporal speech-event annotations (start/end times and total duration), obtained with the alignment procedure described by~\citet{lopez2022iterative}. A portion of OK Aura was released publicly as part of the Albayzin 2024 Wake-up Word Detection Challenge~\citelanguageresource{ok_aura_zenodo} \citep{lopez2024albayzin}.

\subsubsection{Demographic groups}
\label{sec:demographic_groups}

For bias assessment, we consider three demographic attributes available in OK Aura: sex, age, and Spanish accent variety. We perform univariate analyses, evaluating each attribute independently. Sex is treated as a binary variable (Female/Male); age is grouped into 0--20, 21--30, 31--40, 41--50, and 51+; and accent labels cover the full annotation set: Unknown, Central Southern Spain, Southern Spain, Caribbean, Northern Spain, Northwestern Spain, Chilean, Eastern and Balearic Spain, Non-Native, Rio Plata, Canary Islands, Central America, Andean Pacific, Mexico, and Philippines.

\subsubsection{Train and validation splits}
\label{sec:demographics_train_validation}

\begin{figure}[h!]
    \centering
    \includegraphics[width=1.0\linewidth]{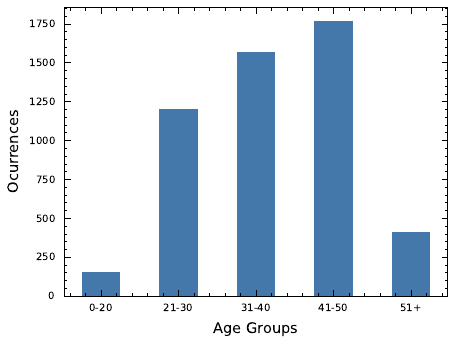}
    \caption{Age distribution in the OK Aura Database (training and validation).}
    \label{fig:age_group_distribution}
\end{figure}

\begin{figure}[h!]
    \centering
    \includegraphics[
        width=\columnwidth,
        trim=0.1cm 0.0cm 0cm 0cm,
        clip
    ]{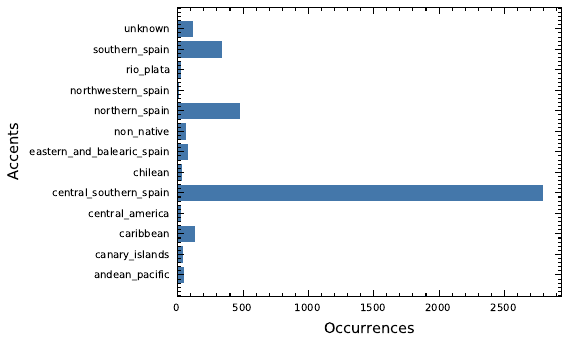}
    \caption{Accent distribution in the OK Aura Database (training and validation).}
    \label{fig:accent_distribution}
\end{figure}

We next analyze demographic distributions in the OK Aura training and validation splits to characterize representation imbalances. Table~\ref{tab:ok_aura_sex} reports the sex distribution, indicating a higher proportion of Male than Female samples. The age distribution has an average speaker age of 37 years, with most samples concentrated between 20 and 50 years old, and comparatively few samples from speakers under 20 or over 51 (Figure~\ref{fig:age_group_distribution}). Finally, accent labels are highly skewed, with Central Southern Spain dominating the training/validation data (Figure~\ref{fig:accent_distribution}).

\begin{table}[h]
    \centering
    \begin{tabular}{l c c}
    \hline
    \textbf{Sex} & \textbf{\# Samples} & \textbf{Percentage}\\
    \hline
    Female & 2131 & 41.74\% \\
    Male & 2974 & 58.26\% \\
    \hline
    \end{tabular}
    \caption{Number of samples by Sex in the OK Aura Database (training and validation).}
    \label{tab:ok_aura_sex}
\end{table}

\subsubsection{Test split}
\label{sec:demographics_test}

The OK Aura test split contains 575 samples of 47 unique speaker. Sex remains imbalanced (Table~\ref{tab:test_ok_aura_sex}), and the age distribution is uneven, with a strong overrepresentation of middle-aged adults (41--50) and no samples from speakers aged 0--20 (Table~\ref{tab:test_ok_aura_age_groups}). Accent coverage in the test set is also limited (Table~\ref{tab:test_ok_aura_accent_groups}), with several regional varieties either absent or severely underrepresented.

\begin{table}[b!]
    \centering
    \begin{tabular}{l c c}
    \hline
    \textbf{Sex} & \textbf{\# Samples} & \textbf{\# Speakers}\\
    \hline
    Female & 254 (44.88\%) & 20 \\
    Male & 321 (55.12\%) & 27 \\
    \hline
    \end{tabular}
    \caption{Number of samples by Sex in the OK Aura Database (test).}
    \label{tab:test_ok_aura_sex}
\end{table}

To ensure reliable subgroup estimates, we exclude demographic groups with fewer than 20 test samples from bias quantification and fairness reporting. Furthermore, because some of the retained groups still feature a limited number of unique speakers (e.g., only 20 female speakers), subsequent fairness metrics should be interpreted as indicative trends rather than absolute guarantees of generalizability.

\begin{table}[h]
    \centering
    \begin{tabular}{l c c}
    \hline
    \textbf{Age Group} & \textbf{\# Samples} & \textbf{\# Speakers} \\
    \hline
    0-20 & 0  & 0 \\
    21-30 & 135 & 11 \\
    31-40 & 138 & 11 \\
    41-50 & 295 & 24 \\
    51+ & 7 & 1\\
    \hline
    \end{tabular}
    \caption{Number of samples by Age Group in the OK Aura Database (test).}
    \label{tab:test_ok_aura_age_groups}
\end{table}

\begin{table}[h]
    \centering
    \begin{tabular}{l c c}
    \hline
    \textbf{Accent} & \textbf{\# Samples} & \textbf{\# Speakers}\\
    \hline
    central southern & 313 & 26\\
    eastern \& balearic & 15 & 1\\
    non native & 49 & 4\\
    northern & 90 & 7\\
    southern & 84 & 7\\
    unknown & 12 & 2\\
    \hline
    \end{tabular}
    \caption{Number of samples and speakers by regional Spanish accent in the OK Aura Database (test).}
    \label{tab:test_ok_aura_accent_groups}
\end{table}

\section{Experimental Setup}
\label{sec:experiments}
We first describe the WuW model, which is designed for on-device inference (Section~\ref{sec:sgru}) and the training procedure (Section~\ref{sec:training_setup}). In the same section, we also detail how data augmentation and knowledge distillation are integrated into training. Finally, we define the metrics used to quantify data imbalance and predictive disparities (Section~\ref{sec:atomic_evaluation}).

\subsection{WuW detection model}
\label{sec:sgru}

We adopt a Gated Recurrent Unit (GRU)-based classifier as a practical trade-off between real-time efficiency and WuW detection accuracy. The model consists of a single GRU layer with 200 hidden units followed by a fully connected classification layer that discriminates between \textit{WuW} and \textit{unknown}. We refer to this architecture as \texttt{device-sgru}; it counts 145.6k parameters, and one inference takes $\sim$25~ms on a Pixel XL device~\citep{lopez2023robust}, making it suitable for real-time inference on-device.

The input features are 13 MFCCs extracted with a 100~ms analysis window and a 50~ms hop, yielding 29 frames for a 1.5~s audio window. We replace the zeroth MFCC coefficient with the log-energy to better capture overall signal intensity~\citep{lopez2023robust}.

For clarification, the \texttt{w2v-BERT2-kws} model described in Section~\ref{sec:ssl_models} is just used to distill knowledge from it, it is not intended to be executed or deployed on device. 

\subsection{Training}
\label{sec:training_setup}
All hyperparameters were set based on our previous research \citep{lopez2023robust}, prioritizing preservation of the strong baseline detection performance. Specifically, models are trained for up to 700 epochs by minimizing cross-entropy loss with a batch size of 128. We use Adam with an initial learning rate (LR) of 0.001 and reduce the LR by a factor of 10 when validation performance plateaus; training stops after four successive LR reductions without improvement.

To improve robustness under diverse acoustic conditions, we apply additive noise and reverberation (RIR convolution) during validation. Because such augmentation introduces additional variance in the validation loss, we select the final checkpoint as the one minimizing the mean of the three lowest validation-loss values across epochs, which stabilizes model selection under stochastic validation augmentation.

This procedure is used to train both the primary \texttt{device-sgru} model and the SSL-based teacher \texttt{w2v-BERT2-kws} model used for KD. The \texttt{device-sgru} model is trained from scratch with uniformly initialized weights; we refer to this configuration as \texttt{baseline}. For \texttt{w2v-BERT2-kws}, the w2v-BERT~2.0 pre-trained encoder is kept frozen, and only the task-specific layers are trained from scratch (uniform initialization).

We then integrate two demographics-unaware training strategies for bias mitigation: data augmentation and KD. During training, augmentations are applied with probability $p=0.2$ (i.e., 20\% of training samples), aiming to preserve strong baseline characteristics while injecting robustness-inducing perturbations.

The following augmentation configuration is used:
\begin{itemize}
    \item \textbf{FilterAugment:} number of frequency bands uniformly sampled in $[3,9]$, minimum bandwidth 187~Hz, gain sampled in $\pm6$~dB.
    \item \textbf{FreqMixStyle:} $\alpha=0.4$ for the Beta distribution; mixing is restricted to pairs of samples with the same label.
    \item \textbf{Frequency masking:} $W_F=30$ and $\nu=128$ mel channels.
\end{itemize}

For KD, we initialize the student \texttt{device-sgru} with the weights of the pre-trained \texttt{baseline} to accelerate convergence. We then optimize the distillation objective in Eq.~\ref{eq:distillation}. During distillation we switch to Stochastic Gradient Descent (SGD) (momentum 0.9, weight decay $10^{-4}$), as it can yield flatter minima and improved generalization. The initial LR is 0.0001 with an on-plateau scheduler, and we set $\delta=0.2$ and $\tau=2$.

\subsection{Evaluation and metrics for bias quantification}
\label{sec:atomic_evaluation}

Predictive disparities across demographic groups are often linked to data imbalance, where underrepresented cohorts tend to suffer degraded performance~\citep{barocas2016big}. We therefore relate demographic imbalance in the OK Aura training/validation data to disparities observed in model predictions on the test split. Concretely, we quantify imbalance in the input data  (Section~\ref{subsec:bias_in_data}) and quantify predictive differences (Section~\ref{subsec:bias_in_pred}).

\subsubsection{Bias in data}
\label{subsec:bias_in_data}

To quantify demographic imbalance in the dataset, we use Disparate Impact (DI), a commonly used ratio-based metric in algorithmic fairness. Let $G$ denote the set of demographic groups, with $a,d\in G$ representing an advantaged and disadvantaged group, and let $Y\in\{0,1\}$ denote the binary label ($1$ for WuW presence and $0$ otherwise). DI is defined as:
\begin{equation}
    DI = \frac{P(Y=1 \mid G=d)}{P(Y=1 \mid G=a)} .
\end{equation}
For multi-valued attributes (e.g., accent, age groups), we report the maximum ratio (or equivalently the most imbalanced pair) across all group pairs.

\begin{table*}[t]
\centering
\begin{tabular}{l l l c}
\hline
\textbf{Attribute} & \textbf{Advantaged Group} & \textbf{Disadvantaged Group} & \textbf{Disparate Impact} \\
\hline
Sex & Male & Female & 0.7170 \\
Age & 41-50 & 21-30 &  0.6804 \\
Accent & central\_southern & northern & 0.1692 \\
\hline
\end{tabular}
\caption{Disparate Impact by attribute in train and validation splits of OK Aura database.}
\label{tab:data_di}
\end{table*}

\subsubsection{Bias in predictions}
\label{subsec:bias_in_pred}

We follow the pairwise group comparison protocol described by ~\citet{singh2023brief} to assess predictive disparities. We evaluate WuW detection on fixed 1.5~s windows and use a fixed decision threshold of 0.5. We report performance using the F1-score:
\begin{equation}
    \mathrm{F1} = 2 \times \frac{\mathrm{Precision}\times \mathrm{Recall}}{\mathrm{Precision}+\mathrm{Recall}},
\end{equation}
where
\begin{equation}
    \mathrm{Precision} = \frac{TP}{TP+FP}
\end{equation}

\begin{equation}
    \mathrm{Recall} = \frac{TP}{TP+FN}
\end{equation}

\textbf{Predictive Disparity (PD).}
We define PD as the maximum absolute difference in F1-score across demographic groups:
\begin{equation}
    PD = \max_{i,j \in G}\left| \mathrm{F1}(g_i) - \mathrm{F1}(g_j) \right|,
\end{equation}
where $g_i,g_j\in G$ denote group identities. Larger values indicate stronger performance gaps and potential fairness concerns.

\textbf{Relative reduction of predictive disparity (RRPD).}
To compare mitigation strategies, we report the relative reduction in PD with respect to the baseline:
\begin{equation}
    RRPD = 100 \times \frac{PD_{\mathrm{baseline}} - PD_{\mathrm{technique}}}{PD_{\mathrm{baseline}}}.
\end{equation}
Here, $PD_{\mathrm{baseline}}$ denotes disparity for the baseline model and $PD_{\mathrm{technique}}$ for the model trained with a given mitigation technique.

\section{Results and Discussion}
\label{sec:results_discussion}

We report (i) demographic imbalance in the OK Aura training/validation splits (data bias; Section~\ref{sec:bias_in_data}), (ii) predictive disparities of the \texttt{baseline} WuW detector on the OK Aura test split (prediction bias; Section~\ref{sec:bias_in_pred}), and (iii) the impact of demographics-agnostic training strategies for mitigation (Section~\ref{sec:bias_mitigation}). Following our evaluation protocol, demographic groups with fewer than 20 test samples are excluded from bias quantification to ensure stable subgroup estimates.

\begin{table}[h]
\centering
\begin{tabular}{l c c}
\toprule
\textbf{Group} & \textbf{F1-score} & \textbf{Support} \\
\midrule
Male   & \textbf{0.9863} & 296 \\
Female & 0.9825          & 204 \\

\midrule

21--30 & \textbf{0.9956} & 115 \\
31--40 & 0.9828          & 118 \\
41--50 & 0.9827          & 265 \\
\midrule
southern\_spain          & 0.9818          & 84  \\
central\_southern\_spain & \textbf{0.9873} & 278 \\
northern\_spain          & 0.9781          & 70  \\
non\_native              & 0.9870          & 39  \\
\midrule
\textbf{PD (sex)}    & \multicolumn{2}{c}{0.0038} \\
\textbf{PD (age)}    & \multicolumn{2}{c}{0.0129} \\
\textbf{PD (accent)} & \multicolumn{2}{c}{0.0092} \\
\bottomrule
\end{tabular}
\caption{Performance across demographic groups with predictive disparity (PD) for baseline model.}
\label{tab:bias_pred_baseline}
\end{table}

\subsection{Bias quantification}
\label{sec:bias_quantification}

\subsubsection{Bias in data}
\label{sec:bias_in_data}

Table~\ref{tab:data_di} reports DI for the OK Aura training/validation splits, revealing systematic representation imbalances across all examined attributes. Male speakers are overrepresented relative to female speakers ($DI=0.717$), the 41--50 age group dominates compared to 21--30 ($DI=0.6804$), and accent imbalance is most severe: Central Southern Spain is disproportionately represented relative to Northern Spain ($DI=0.1692$). These skewed distributions are likely to affect generalization and may translate into unequal predictive performance across groups.

\subsubsection{Bias in predictions}
\label{sec:bias_in_pred}

Table \ref{tab:bias_pred_baseline} presents subgroup F1-scores and PD for the \texttt{baseline} model. Sex-related disparity is small but measurable ($PD=0.0038$), with slightly higher F1 for male speakers (0.9863 vs.\ 0.9825). Age exhibits the largest performance gap ($PD=0.0129$): the 21--30 group performs best (0.9956), while the 41--50 group performs worst (0.9827), highlighting older adults as a key cohort for mitigation. Accent disparities are also evident ($PD=0.0092$), where Central Southern Spain achieves the highest F1 (0.9873) and Northern Spain is lower (0.9781), suggesting that accent variability remains a relevant source of error.

Overall, the baseline results indicate that demographic imbalance in the training data co-occurs with predictive disparities, motivating mitigation methods that increase robustness without requiring demographic labels.

\subsection{Bias mitigation and analysis}
\label{sec:bias_mitigation}
\begin{table*}[t]
\centering
\begin{tabular}{l c c c}
\hline
\textbf{Classifier} & \textbf{Sex RRPD (\%)} & \textbf{Age RRPD (\%)} & \textbf{Accent RRPD (\%)} \\
\hline
w2v-BERT2-kws & 79.64 & \textbf{85.35} & 41.05 \\
\hline
\end{tabular}
\caption{w2v-BERT2-kws Relative Reduction of Predictive Disparity for demographic attributes in comparison to baseline model. It demonstrates reduced PD across sex, age, and accent categories.}
\label{tab:ssl_gap_reduction}
\end{table*}

\begin{table*}[t]
\centering
\begin{tabular}{l r r r}
\hline
\textbf{Experimental Setting} & \textbf{Sex RRPD (\%)} & \textbf{Age RRPD (\%)} & \textbf{Accent RRPD (\%)} \\
\hline
DIR & 67.35 & 0.00 & -20.13 \\
FreqMixStyle & -21.42 & 34.12 & \textbf{40.48} \\
FilterAugment & \textbf{88.26} & 30.14 & -40.19 \\
FreqMasking & 39.94 & \textbf{83.65} & \textbf{40.48} \\
KD & 67.35 & 15.10 & -20.13 \\
KD + FreqMasking & 21.24 & 15.10 & -40.19 \\
\hline
\end{tabular}
\caption{Relative Reduction of Predictive Disparity (RRPD) across sex, age, and accent for different training techniques. Higher is better (negative values indicate increased disparity).}
\label{tab:rrpd_all}
\end{table*}

First, the high-capacity SSL-based classifier \texttt{w2v-BERT2-kws} exhibits substantially lower disparities than \texttt{baseline} (Table~\ref{tab:ssl_gap_reduction}), indicating that large-scale SSL pretraining can reduce, but not eliminate, demographic performance gaps. This motivates its use as a teacher model for KD. 

Table~\ref{tab:rrpd_all} reports the relative reduction of predictive disparity achieved by each technique with respect to \texttt{baseline}. We observe that augmentation and KD show attribute-dependent behavior. Specifically, DIR only improves disparity for sex (67.35\% RRPD), suggesting that device-specific impulse responses may not capture the heterogeneous acoustic variations typically associated with demographic attributes. FilterAugment yields the largest reduction in sex disparity (88.26\% RRPD) and also improves age fairness (30.14\% RRPD), but it increases accent disparity. This suggests that while smooth frequency-energy perturbations help reduce reliance on certain demographic-specific spectral cues, they do not universally benefit all attributes. On the other hand, FreqMixStyle improves age and accent fairness but degrades sex fairness (negative RRPD), indicating that frequency-wise statistics mixing affects demographic attributes differently and may not generalize across all cues simultaneously. We hypothesize that FreqMixStyle and FilterAugment, underperform on some speaker demographics as they can reshape frequency statistics too aggressively. They may destroy critical formant/prosodic cues and yielding mixed fairness gains at higher error cost.


\begin{table}[b]
\centering
\begin{tabular}{l c c}
\hline
\textbf{Group} & \textbf{F1-score} & \textbf{Support} \\
\hline
Male & 0.9828 & 296 \\
Female & \textbf{0.9851} & 204 \\
\hline
21--30 & \textbf{0.9880} & 115 \\
31--40 & 0.9828 & 118 \\
41--50 & 0.9847 & 265 \\
\hline
southern\_spain & 0.9818 & 84 \\
central\_southern\_spain & 0.9835 & 278 \\
northern\_spain & 0.9781 & 70 \\
non\_native & \textbf{0.9870} & 39 \\
\hline
\textbf{PD (sex)} & \multicolumn{2}{c}{0.0023} \\
\textbf{PD (age)} & \multicolumn{2}{c}{0.0052} \\
\textbf{PD (accent)} & \multicolumn{2}{c}{0.0089} \\
\hline
\end{tabular}
\caption{Predictive Disparity using FreqMasking technique to train device-sgru model.}
\label{tab:pd_freqmasking}
\end{table}

In contrast, Frequency Masking provides the most consistent gains across all attributes, achieving a strong reduction in age disparity (83.65\%) while also narrowing the gaps for sex and accent. Furthermore, Table~\ref{tab:pd_freqmasking} shows that these gains are achieved while maintaining competitive subgroup F1-scores, making Frequency Masking a suitable fairness‑oriented augmentation. Suppressing specific frequency bands appears to force the model to distribute evidence across multiple regions of the spectrum rather than overfitting to a single demographic‑correlated band (e.g., the F0 or low‑formant region). This distributed attention both improves overall robustness and reduces reliance on demographic‑specific cues.


Finally, KD reduces sex and age disparity but does not consistently reduce accent disparity. One plausible explanation is that accent invariance is constrained by limited accent diversity in the labeled in-domain data used during distillation, which may limit the teacher's ability to provide accent-neutral soft targets. Finally, combining KD with Frequency Masking does not improve over the best single-technique settings and can degrade results for some attributes, suggesting an interaction between stochastic spectral corruption and logit matching that may be difficult for a small student architecture to optimize jointly.

\section{Conclusion and Future Work}
\label{sec:conclusion}

This work shows that demographic-agnostic training can mitigate bias in Wake-up Word detection without requiring demographic labels during training. We studied two complementary families of methods: (i) data augmentation that perturbs or removes frequency information, and (ii) knowledge distillation (KD) from a large pre-trained speech model. Across speaker groups defined by sex, age, and accent, these approaches reduce performance disparities while maintaining competitive overall accuracy.


Our results highlight that augmentation design is critical for effective mitigation. In particular, frequency-energy perturbations and statistics mixing exhibited attribute-dependent behavior, sometimes degrading fairness for specific groups. Contrary, Frequency Masking emerged as the most robust single technique. It consistently achieved a relative reduction in predictive disparity (RRPD) of 39.94\% (sex), 83.65\% (age), and 40.48\% (accent). By suppressing specific frequency bands, Frequency Masking prevents the model from overfitting to demographic-correlated acoustic cues (e.g., fundamental frequency) and forces it to distribute evidence across the broader spectrum. Additionally, KD achieved high RRPD, especially for sex and age, but showed limited impact on accent. This suggests that accent-invariant transfer remains constrained by the limited accent diversity available in the in-domain distillation data.


Future work will extend this analysis to intersectional fairness settings (e.g., older females with specific regional accents) and broaden demographic coverage by curating more balanced data across attributes. Particular emphasis will be placed on underrepresented age and accent groups.

\paragraph{Limitations.}
(i) Our analysis is univariate and does not capture intersectional effects (e.g., older female speakers with regional accents). (ii) The training/validation procedure includes out-of-domain audio sources, which can introduce distribution mismatch and metadata inconsistencies. (iii) Several demographic groups are underrepresented in the test split and were excluded from fairness reporting; furthermore, the limited number of speakers within the retained groups implies that our specific conclusions regarding them should be interpreted with caution. (iv)~F1 does not decompose disparities into false accepts and false rejects, which carry asymmetric costs in WuW detection. (v)~Conclusions at a fixed threshold of 0.5 may not generalize across operating points. Future work should adopt multi-objective evaluation frameworks that jointly optimize overall accuracy, per-group F1, and cross-attribute fairness, rather than treating each in isolation.

\paragraph{Ethics Statement}
This research addresses fairness in speech systems, which has positive ethical implications for reducing discrimination. The OK Aura dataset involves anonymized speakers with informed consent. Our demographics-agnostic approach specifically avoids requiring sensitive demographic labels during deployment, protecting user privacy. However, we acknowledge that bias mitigation techniques may have unintended effects on other demographic groups not examined in this study, and that univariate analysis may miss intersectional discrimination patterns.

\section{Acknowledgments}
This project has been partially funded by the Spanish Project 6G-RIEMANN (Grant Agreement No. 2022/0005420) and by the European Union’s Horizon 2020 RIA ELOQUENCE project (Grant Agreement No. 101135916). Views and opinions expressed are, however, those of the author(s) only and do not necessarily reflect those of the European Union or European Commission-EU. Neither the European Union nor the granting authority can be held responsible for them.

\section{Bibliographical References}
\label{sec:reference}
\bibliographystyle{lrec2026-natbib}

\begin{thebibliography}{48}
\expandafter\ifx\csname natexlab\endcsname\relax\def\natexlab#1{#1}\fi

\bibitem[{Attanasio et~al.(2024)Attanasio, Savoldi, Fucci, and Hovy}]{attanasio2024twists}
Giuseppe Attanasio, Beatrice Savoldi, Dennis Fucci, and Dirk Hovy. 2024.
\newblock \href {https://doi.org/10.18653/v1/2024.emnlp-main.1188} {Twists, humps, and pebbles: Multilingual speech recognition models exhibit gender performance gaps}.
\newblock In \emph{Proceedings of the 2024 Conference on Empirical Methods in Natural Language Processing}, pages 21318--21340, Miami, Florida, USA. Association for Computational Linguistics.

\bibitem[{Bailey and Plumbley(2021)}]{bailey2021gender}
Andrew Bailey and Mark~D Plumbley. 2021.
\newblock \href {https://ieeexplore.ieee.org/stamp/stamp.jsp?arnumber=9615933&casa_token=FItrcAm19_wAAAAA:QrsfabjFUNJMf8FePf0YQv5oms5oHKvtmZiOtQwBhCm1XGc9NZvVIPl69OHGaPWPXmq6-jyCsQ&tag=1} {Gender bias in depression detection using audio features}.
\newblock In \emph{IEEE 29th European Signal Processing Conference (EUSIPCO)}, pages 596--600.

\bibitem[{Barocas and Selbst(2016)}]{barocas2016big}
Solon Barocas and Andrew~D. Selbst. 2016.
\newblock \href {https://doi.org/10.15779/Z38BG31} {Big data’s disparate impact}.
\newblock \emph{California Law Review}, 104(3):671--732.

\bibitem[{Barrault et~al.(2023)Barrault, Chung, Meglioli, Dale, Dong, Duppenthaler, Duquenne, Ellis, Elsahar, Haaheim et~al.}]{barrault2023seamless}
Lo{\"\i}c Barrault, Yu-An Chung, Mariano~Coria Meglioli, David Dale, Ning Dong, Mark Duppenthaler, Paul-Ambroise Duquenne, Brian Ellis, Hady Elsahar, Justin Haaheim, et~al. 2023.
\newblock \href {https://arxiv.org/abs/2312.05187} {Seamless: Multilingual expressive and streaming speech translation}.
\newblock \emph{arXiv preprint arXiv:2312.05187}.

\bibitem[{Chai et~al.(2022)Chai, Wang, Chen, He, Song, and Li}]{chai2022fairness}
Junyi Chai, Zhihao Wang, Jiaxin Chen, Hao He, Dawn Song, and Xia Li. 2022.
\newblock \href {https://papers.neurips.cc/paper_files/paper/2022/file/79dc391a2c1067e9ac2b764e31a60377-Paper-Conference.pdf} {Fairness without demographics through knowledge distillation}.
\newblock In \emph{Advances in Neural Information Processing Systems (NeurIPS)}, volume~35.

\bibitem[{Chen et~al.(2022)Chen, Chen, Wu, Qian, Wang, Liu, Qian, and Zeng}]{chen2022large}
Zhengyang Chen, Sanyuan Chen, Yu~Wu, Yao Qian, Chengyi Wang, Shujie Liu, Yanmin Qian, and Michael Zeng. 2022.
\newblock \href {https://doi.org/10.1109/ICASSP43922.2022.9747814} {Large-scale self-supervised speech representation learning for automatic speaker verification}.
\newblock In \emph{IEEE International Conference on Acoustics, Speech and Signal Processing (ICASSP)}, pages 6147--6151.

\bibitem[{Choi and Choi(2025)}]{Choi_Choi_2025}
Anna Seo~Gyeong Choi and Hoon Choi. 2025.
\newblock \href {https://doi.org/10.1609/aies.v8i1.36574} {Fairness of automatic speech recognition: Looking through a philosophical lens}.
\newblock \emph{Proceedings of the AAAI/ACM Conference on AI, Ethics, and Society}, 8(1):605--614.

\bibitem[{Dheram et~al.(2022)Dheram, Ramakrishnan, Raju, Chen, King, Powell, Saboowala, Shetty, and Stolcke}]{dheram2022toward}
Pranav Dheram, Murugesan Ramakrishnan, Anirudh Raju, I-Fan Chen, Brian King, Katherine Powell, Melissa Saboowala, Karan Shetty, and Andreas Stolcke. 2022.
\newblock \href {https://www.academia.edu/117404158/Toward_Fairness_in_Speech_Recognition_Discovery_and_mitigation_of_performance_disparities} {Toward fairness in speech recognition: Discovery and mitigation of performance disparities}.
\newblock \emph{INTERSPEECH}.

\bibitem[{ElGhazaly et~al.(2025)ElGhazaly, Mirheidari, Christensen, and Moosavi}]{elghazaly2025fairasr}
Hend ElGhazaly, Bahman Mirheidari, Heidi Christensen, and Nafise~Sadat Moosavi. 2025.
\newblock \href {https://doi.org/10.18653/v1/2025.findings-emnlp.1044} {Fairness in automatic speech recognition isn{'}t a one-size-fits-all}.
\newblock In \emph{Findings of the Association for Computational Linguistics: EMNLP}, pages 19169--19178, Suzhou, China.

\bibitem[{Feng et~al.(2021)Feng, Kudina, Halpern, and Scharenborg}]{feng2021quantifying}
Siyuan Feng, Olya Kudina, Bence~Mark Halpern, and Odette Scharenborg. 2021.
\newblock \href {https://arxiv.org/abs/2103.15122} {Quantifying bias in automatic speech recognition}.
\newblock \emph{arXiv preprint arXiv:2103.15122}.

\bibitem[{Fuckner et~al.(2023)Fuckner, Horsman, Wiggers, and Janssen}]{fuckner2023uncovering}
Marcio Fuckner, Sophie Horsman, Pascal Wiggers, and Iskaj Janssen. 2023.
\newblock \href {https://scholar.google.com/scholar?hl=en&as_sdt=0%2C5&q=Uncovering+bias+in+asr+systems%3A+Evaluating+wav2vec2+and+whisper+for+dutch+speakers&btnG=} {Uncovering bias in asr systems: Evaluating wav2vec2 and whisper for dutch speakers}.
\newblock In \emph{IEEE International Conference on Speech Technology and Human-Computer Dialogue (SpeD)}, pages 146--151.

\bibitem[{Garg et~al.(2018)Garg, Schiebinger, Jurafsky, and Zou}]{garg2018word}
Nikhil Garg, Londa Schiebinger, Dan Jurafsky, and James Zou. 2018.
\newblock \href {https://www.pnas.org/doi/abs/10.1073/pnas.1720347115} {Word embeddings quantify 100 years of gender and ethnic stereotypes}.
\newblock \emph{Proceedings of the National Academy of Sciences}, 115(16):E3635--E3644.

\bibitem[{Han et~al.(2025)Han, Landini, Rohdin, Silnova, Diez, and Burget}]{han2025leveraging}
Jiangyu Han, Federico Landini, Johan Rohdin, Anna Silnova, Mireia Diez, and Luk{\'a}{\v{s}} Burget. 2025.
\newblock \href {https://ieeexplore.ieee.org/abstract/document/10889475?casa_token=W0dHhapj0TEAAAAA:juhneHrbIk-SxU7Rjw3DUCLY6jtZzgUjoRnqwvSuQeAZgTUmDQwUN4BkW0fiqQug2DDrF4W0Mg} {Leveraging self-supervised learning for speaker diarization}.
\newblock In \emph{IEEE International Conference on Acoustics, Speech and Signal Processing (ICASSP)}, pages 1--5.

\bibitem[{Harnsberger et~al.(2008)Harnsberger, Shrivastav, Brown~Jr, Rothman, and Hollien}]{harnsberger2008speaking}
James~D Harnsberger, Rahul Shrivastav, William~S Brown~Jr, Howard Rothman, and Harry Hollien. 2008.
\newblock \href {https://www.sciencedirect.com/science/article/pii/S0892199706000889?casa_token=HlYDdZn8ymIAAAAA:VlLTpjdn1f9Fp8xIBULMJrVqL4bcEvUysA4x3PVX-VQnjmVZRxWd5sSFOj0iTKdMDsB9o4u6K0U} {Speaking rate and fundamental frequency as speech cues to perceived age}.
\newblock \emph{Journal of voice}, 22(1):58--69.

\bibitem[{Harris et~al.(2024)Harris, Mgbahurike, Kumar, and Yang}]{harris2024modeling}
Camille Harris, Chijioke Mgbahurike, Neha Kumar, and Diyi Yang. 2024.
\newblock \href {https://aclanthology.org/2024.findings-emnlp.890/} {Modeling gender and dialect bias in automatic speech recognition}.
\newblock In \emph{Findings of the Association for Computational Linguistics: EMNLP}, pages 15166--15184.

\bibitem[{Hutiri et~al.(2023)Hutiri, Ding, Kawsar, and Mathur}]{hutiri2023tiny}
Wiebke Hutiri, Aaron~Yi Ding, Fahim Kawsar, and Akhil Mathur. 2023.
\newblock \href {https://dl.acm.org/doi/abs/10.1145/3591867} {Tiny, always-on, and fragile: Bias propagation through design choices in on-device machine learning workflows}.
\newblock \emph{ACM Transactions on Software Engineering and Methodology}, 32(6):1--37.

\bibitem[{K{\"e}puska and Breitfeller(2006)}]{10.1117/12.666025}
Veton K{\"e}puska and Jason Breitfeller. 2006.
\newblock \href {https://doi.org/10.1117/12.666025} {{Wake-up-word speech recognition application for first responder communication enhancement}}.
\newblock In \emph{Sensors, and Command, Control, Communications, and Intelligence (C3I) Technologies for Homeland Security and Homeland Defense V}, volume 6201, page 62011E. International Society for Optics and Photonics, SPIE.

\bibitem[{Kim et~al.(2022)Kim, Yang, Kim, Park, Lee, and Chang}]{kim2022domain}
Byeonggeun Kim, Seunghan Yang, Jangho Kim, Hyunsin Park, Juntae Lee, and Simyung Chang. 2022.
\newblock \href {https://arxiv.org/abs/2206.12513} {Domain generalization with relaxed instance frequency-wise normalization for multi-device acoustic scene classification}.
\newblock \emph{arXiv preprint arXiv:2206.12513}.

\bibitem[{Kim et~al.(2021)Kim, Han, and Ko}]{kim2021specmix}
Gwantae Kim, David~K Han, and Hanseok Ko. 2021.
\newblock \href {https://arxiv.org/abs/2108.03020} {Specmix: A mixed sample data augmentation method for training withtime-frequency domain features}.
\newblock \emph{arXiv preprint arXiv:2108.03020}.

\bibitem[{Koudounas et~al.(2024)Koudounas, Giobergia, Pastor, and Baralis}]{koudounas2024contrastive}
Alkis Koudounas, Flavio Giobergia, Eliana Pastor, and Elena Baralis. 2024.
\newblock \href {https://arxiv.org/abs/2406.14686} {A contrastive learning approach to mitigate bias in speech models}.
\newblock \emph{arXiv preprint arXiv:2406.14686}.

\bibitem[{Këpuska(2011)}]{Kepuska11}
Veton Këpuska. 2011.
\newblock \href {https://doi.org/10.5772/16242} {Wake-up-word speech recognition}.
\newblock In Ivo Ipsic, editor, \emph{Speech Technologies}, chapter~12. IntechOpen, London.

\bibitem[{Labrador et~al.(2025)Labrador, Zhu, Zhao, Scarpati, Wang, Lozano-Diez, and Lopez-Moreno}]{zhao2025personalizingkws}
Beltrán Labrador, Pai Zhu, Guanlong Zhao, Angelo~Scorza Scarpati, Quan Wang, Alicia Lozano-Diez, and Ignacio Lopez-Moreno. 2025.
\newblock \href {https://doi.org/10.1109/ICASSP49660.2025.10890730} {Personalizing keyword spotting with speaker information}.
\newblock In \emph{IEEE International Conference on Acoustics, Speech and Signal Processing (ICASSP)}, pages 1--5.

\bibitem[{Lin et~al.(2024)Lin, Lin, Lin, Liu, and Lee}]{lin2024sslbias}
Yi-Cheng Lin, Tzu-Quan Lin, Hsi-Che Lin, Andy~T Liu, and Hung-yi Lee. 2024.
\newblock \href {https://doi.org/10.21437/INTERSPEECH.2024-454} {On the social bias of speech self-supervised models}.
\newblock In \emph{Proceedings of INTERSPEECH}, pages 4638--4642.

\bibitem[{L{\'o}pez and Luque(2024)}]{lopez2024albayzin}
Fernando L{\'o}pez and Jordi Luque. 2024.
\newblock \href {https://catedrartve.unizar.es/albayzin2024.html} {Albayzin evaluation 2024: Wake-up word detection challenge}.

\bibitem[{L{\'o}pez-Espejo et~al.(2021)L{\'o}pez-Espejo, Tan, Hansen, and Jensen}]{lopez2021deep}
Iv{\'a}n L{\'o}pez-Espejo, Zheng-Hua Tan, John~HL Hansen, and Jesper Jensen. 2021.
\newblock \href {https://ieeexplore.ieee.org/abstract/document/9665775/} {Deep spoken keyword spotting: An overview}.
\newblock \emph{IEEE Access}, 10:4169--4199.

\bibitem[{López and Luque(2022)}]{lopez2022iterative}
Fernando López and Jordi Luque. 2022.
\newblock \href {https://doi.org/10.21437/IberSPEECH.2022-10} {{Iterative pseudo-forced alignment by acoustic CTC loss for self-supervised ASR domain adaptation}}.
\newblock In \emph{Proceedings of IberSPEECH}, pages 46--50.

\bibitem[{López et~al.(2023)López, Luque, Segura, and Gómez}]{lopez2023robust}
Fernando López, Jordi Luque, Carlos Segura, and Pablo Gómez. 2023.
\newblock \href {https://arxiv.org/abs/2310.11379} {Robust wake-up word detection by two-stage multi-resolution ensembles}.
\newblock \emph{arXiv preprint arXiv:2310.11379}.

\bibitem[{Martin and Wright(2023)}]{martin2023bias}
Joshua~L Martin and Kelly~Elizabeth Wright. 2023.
\newblock \href {https://academic.oup.com/applij/article-abstract/44/4/613/6901317} {Bias in automatic speech recognition: The case of african american language}.
\newblock \emph{Applied Linguistics}, 44(4):613--630.

\bibitem[{Meng et~al.(2022)Meng, Chou, Liu, and Lee}]{meng2022dont}
Yen Meng, Yi-Hui Chou, Andy~T. Liu, and Hung-yi Lee. 2022.
\newblock \href {https://doi.org/10.1109/ICASSP43922.2022.9747897} {Don't speak too fast: The impact of data bias on self-supervised speech models}.
\newblock In \emph{IEEE International Conference on Acoustics, Speech and Signal Processing (ICASSP)}, pages 3258--3262.

\bibitem[{Mohamed et~al.(2022)Mohamed, Lee, Borgholt, Havtorn, Edin, Igel, Kirchhoff, Li, Livescu, Maaløe, Sainath, and Watanabe}]{mohamed2022self}
Abdelrahman Mohamed, Hung-yi Lee, Lasse Borgholt, Jakob~D. Havtorn, Joakim Edin, Christian Igel, Katrin Kirchhoff, Shang-Wen Li, Karen Livescu, Lars Maaløe, Tara~N. Sainath, and Shinji Watanabe. 2022.
\newblock \href {https://doi.org/10.1109/JSTSP.2022.3207050} {Self-supervised speech representation learning: A review}.
\newblock \emph{IEEE Journal of Selected Topics in Signal Processing}, 16(6):1179--1210.

\bibitem[{M{\o}rk et~al.(2024)M{\o}rk, Bovbjerg, Kiss, and Tan}]{mork2024noise}
Jacob M{\o}rk, Holger~Severin Bovbjerg, Gergely Kiss, and Zheng-Hua Tan. 2024.
\newblock \href {https://arxiv.org/abs/2403.18560} {Noise-robust keyword spotting through self-supervised pretraining}.
\newblock \emph{arXiv preprint arXiv:2403.18560}.

\bibitem[{Morocutti et~al.(2023)Morocutti, Schmid, Koutini, and Widmer}]{morocutti2023device}
Tobias Morocutti, Florian Schmid, Khaled Koutini, and Gerhard Widmer. 2023.
\newblock \href {https://ieeexplore.ieee.org/abstract/document/10289983/?casa_token=RYzRaQ1FtZ4AAAAA:HI-fCnXppQPRj2grhKkwVuECakO44XxdL2oqCDbV-LsU60kS-PdNQoHaHTcsp1nigHVN52OGTA} {Device-robust acoustic scene classification via impulse response augmentation}.
\newblock In \emph{IEEE 31st European Signal Processing Conference (EUSIPCO)}, pages 176--180. IEEE.

\bibitem[{Mujtaba et~al.(2024)Mujtaba, Mahapatra, Arney, Yaruss, Gerlach-Houck, Herring, and Bin}]{mujtaba2024lost}
Dena Mujtaba, Nihar Mahapatra, Megan Arney, J~Yaruss, Hope Gerlach-Houck, Caryn Herring, and Jia Bin. 2024.
\newblock \href {https://doi.org/10.18653/v1/2024.naacl-long.269} {Lost in transcription: Identifying and quantifying the accuracy biases of automatic speech recognition systems against disfluent speech}.
\newblock In \emph{Proceedings of Conference of the North American Chapter of the Association for Computational Linguistics: Human Language Technologies (Volume 1: Long Papers)}, pages 4795--4809, Mexico City, Mexico.

\bibitem[{Nam et~al.(2022)Nam, Kim, and Park}]{nam2022filteraugment}
Hyeonuk Nam, Seong-Hu Kim, and Yong-Hwa Park. 2022.
\newblock \href {https://doi.org/10.1109/ICASSP43922.2022.9747680} {Filteraugment: An acoustic environmental data augmentation method}.
\newblock In \emph{IEEE International Conference on Acoustics, Speech and Signal Processing (ICASSP)}.

\bibitem[{Park et~al.(2019)Park, Chan, Zhang, Chiu, Zoph, Cubuk, and Le}]{park2019specaugment}
Daniel~S Park, William Chan, Yu~Zhang, Chung-Cheng Chiu, Barret Zoph, Ekin~D Cubuk, and Quoc~V Le. 2019.
\newblock \href {https://arxiv.org/abs/1904.08779} {Specaugment: A simple data augmentation method for automatic speech recognition}.
\newblock \emph{arXiv preprint arXiv:1904.08779}.

\bibitem[{Pasad et~al.(2023)Pasad, Shi, and Livescu}]{pasad2023comparative}
Ankita Pasad, Bowen Shi, and Karen Livescu. 2023.
\newblock \href {https://doi.org/10.1109/ICASSP49357.2023.10096149} {Comparative layer-wise analysis of self-supervised speech models}.
\newblock In \emph{IEEE International Conference on Acoustics, Speech and Signal Processing (ICASSP)}, pages 1--5.

\bibitem[{Pastor et~al.(2021)Pastor, De~Alfaro, and Baralis}]{pastor2021looking}
Eliana Pastor, Luca De~Alfaro, and Elena Baralis. 2021.
\newblock \href {https://dl.acm.org/doi/abs/10.1145/3448016.3457284} {Looking for trouble: Analyzing classifier behavior via pattern divergence}.
\newblock In \emph{Proceedings of the International Conference on Management of Data}, pages 1400--1412.

\bibitem[{Peng et~al.(2025)Peng, Mo{\v{s}}ner, Zhang, Plchot, Stafylakis, Burget, and {\v{C}}ernock{\`y}}]{peng2025mhfa}
Junyi Peng, Ladislav Mo{\v{s}}ner, Lin Zhang, Old{\v{r}}ich Plchot, Themos Stafylakis, Luk{\'a}{\v{s}} Burget, and Jan {\v{C}}ernock{\`y}. 2025.
\newblock \href {https://ieeexplore.ieee.org/abstract/document/10889058?casa_token=FbuK82rQ2osAAAAA:UDowTpmmpQGnsNGxbUKO0tUJ5OhNKSUQT05K_UGA7XNdJ1ZB9T1ROUUVkWUCc0AroRZCNlLUPg} {Ca-mhfa: A context-aware multi-head factorized attentive pooling for ssl-based speaker verification}.
\newblock In \emph{IEEE International Conference on Acoustics, Speech and Signal Processing (ICASSP)}, pages 1--5.

\bibitem[{Piat et~al.(2008)Piat, Fohr, and Illina}]{piat2008foreign}
Marina Piat, Dominique Fohr, and Irina Illina. 2008.
\newblock \href {https://www.isca-archive.org/INTERSPEECH_2008/piat08_INTERSPEECH.pdf} {Foreign accent identification based on prosodic parameters}.
\newblock In \emph{INTERSPEECH}, pages 759--762.

\bibitem[{Radford et~al.(2023)Radford, Kim, Xu, Brockman, McLeavey, and Sutskever}]{radford2023robust}
Alec Radford, Jong~Wook Kim, Tao Xu, Greg Brockman, Christine McLeavey, and Ilya Sutskever. 2023.
\newblock Robust speech recognition via large-scale weak supervision.
\newblock In \emph{International Conference on Machine Learning (ICML)}, volume 202, pages 28492--28518. PMLR.

\bibitem[{{Roncel Díaz} et~al.(2024){Roncel Díaz}, Costa, and Hernando}]{ronceldiaz24on}
Daniel {Roncel Díaz}, Federico Costa, and Javier Hernando. 2024.
\newblock \href {https://doi.org/10.21437/IberSPEECH.2024-31} {On the use of audio to improve dialogue policies}.
\newblock In \emph{IberSPEECH}, pages 151--155.

\bibitem[{Singh et~al.(2023)Singh, Xia, Kim, Pirracchio, Chunara, and Feng}]{singh2023brief}
Harvineet Singh, Fan Xia, Mi-Ok Kim, Romain Pirracchio, Rumi Chunara, and Jean Feng. 2023.
\newblock \href {https://arxiv.org/abs/2312.04745} {A brief tutorial on sample size calculations for fairness audits}.
\newblock \emph{arXiv preprint arXiv:2312.04745}.

\bibitem[{Slaughter et~al.(2023)Slaughter, Greenberg, Schwartz, and Caliskan}]{slaughter2023pre}
Isaac Slaughter, Craig Greenberg, Reva Schwartz, and Aylin Caliskan. 2023.
\newblock \href {https://aclanthology.org/2023.findings-emnlp.602/} {Pre-trained speech processing models contain human-like biases that propagate to speech emotion recognition}.
\newblock pages 8967--8989.

\bibitem[{Vandenberghe et~al.(2023)}]{vandenberghe2023augmentation}
Loes Vandenberghe et~al. 2023.
\newblock \href {http://arxiv.org/abs/2312.15499} {Exploring data augmentation in bias mitigation against non-native-accented speech}.
\newblock \emph{arXiv preprint arXiv:2312.15499}.

\bibitem[{Vorperian et~al.(2019)Vorperian, Kent, Lee, and Bolt}]{vorperian2019corner}
Houri~K Vorperian, Raymond~D Kent, Yen Lee, and Daniel~M Bolt. 2019.
\newblock \href {https://pubs.aip.org/asa/jasa/article-abstract/146/5/3255/993903/Corner-vowels-in-males-and-females-ages-4-to-20?redirectedFrom=fulltext} {Corner vowels in males and females ages 4 to 20 years: Fundamental and f1--f4 formant frequencies}.
\newblock \emph{The Journal of the Acoustical Society of America}, 146(5):3255--3274.

\bibitem[{Wang et~al.(2024)Wang, Delgado, Tak, Jung, Shim, Todisco, Kukanov, Liu, Sahidullah, Kinnunen et~al.}]{wang2024asvspoof}
Xin Wang, H{\'e}ctor Delgado, Hemlata Tak, Jee-weon Jung, Hye-jin Shim, Massimiliano Todisco, Ivan Kukanov, Xuechen Liu, Md~Sahidullah, Tomi Kinnunen, et~al. 2024.
\newblock \href {https://arxiv.org/abs/2408.08739} {Asvspoof 5: Crowdsourced speech data, deepfakes, and adversarial attacks at scale}.
\newblock \emph{arXiv preprint arXiv:2408.08739}.

\bibitem[{Yu et~al.(2023)Yu, Jin, Wan, and Wang}]{yu2023few}
Mingdong Yu, Xiaofeng Jin, Bangxian Wan, and Guirong Wang. 2023.
\newblock \href {https://doi.org/10.1109/CISP-BMEI60920.2023.10373303} {A few-shot speech keyword spotting method based on self-supervise learning}.
\newblock In \emph{16th International Congress on Image and Signal Processing, BioMedical Engineering and Informatics (CISP-BMEI)}, pages 1--5.

\bibitem[{Zolnoori et~al.(2024)Zolnoori, Vergez, Xu, Esmaeili, Zolnour, Anne~Briggs, Scroggins, Hosseini~Ebrahimabad, Noble, Topaz et~al.}]{zolnoori2024decoding}
Maryam Zolnoori, Sasha Vergez, Zidu Xu, Elyas Esmaeili, Ali Zolnour, Krystal Anne~Briggs, Jihye~Kim Scroggins, Seyed~Farid Hosseini~Ebrahimabad, James~M Noble, Maxim Topaz, et~al. 2024.
\newblock \href {https://academic.oup.com/jamiaopen/article/7/4/ooae130/7920671?guestAccessKey=} {Decoding disparities: evaluating automatic speech recognition system performance in transcribing black and white patient verbal communication with nurses in home healthcare}.
\newblock \emph{JAMIA open}, 7(4):ooae130.

\end{thebibliography}

\begin{thebibliography}{8}
\expandafter\ifx\csname natexlab\endcsname\relax\def\natexlab#1{#1}\fi

\bibitem[{Cámbara et~al.(2024)Cámbara, Luque, Bonet, López, Farrús, Gómez, and Segura}]{ok_aura_zenodo}
Cámbara, Guillermo and Luque, Jordi and Bonet, David and López, Fernando and Farrús, Mireia and Gómez, Pablo and Segura, Carlos. 2024.
\newblock \href {https://doi.org/10.5281/zenodo.11082517} {\emph{Okey Aura Wake-up Word Dataset}}.
\newblock Zenodo, 1.1.0.

\bibitem[{Hern{\'a}ndez et~al.(2021)Hern{\'a}ndez, Brookes, and Sena}]{multi_angle_multi_distance_micir_zenodo}
Juan Carlos Franco Hern{\'a}ndez and Tim Brookes and Enzo De Sena. 2021.
\newblock \href {https://doi.org/10.5281/zenodo.4633508} {\emph{Multi-Angle, Multi-Distance Microphone Impulse Response Dataset}}.
\newblock Zenodo, 1.0.0.

\bibitem[{Jahan et~al.(2025)Jahan, Sun, Mazumdar, Fagyal, Thebaud, Villalba, Hasegawa-Johnson, Dehak, and Velazquez}]{jahan2025faist}
Maliha Jahan, Yinglun Sun, Priyam Mazumdar, Zsuzsanna Fagyal, Thomas Thebaud, Jesus Villalba, Mark Hasegawa-Johnson, Najim Dehak, and Laureano~Moro Velazquez. 2025.
\newblock \href {https://doi.org/10.21437/Interspeech.2025-1102} {Faist: A benchmark dataset for fairness in speech technology}.
\newblock In \emph{Proceedings of Interspeech}, pages 1343--1347.

\bibitem[{Joachim~Thiemann and Vincent(2013)}]{demand}
Joachim Thiemann, Nobutaka Ito and Emmanuel Vincent. 2013.
\newblock \href {https://doi.org/10.1121/1.4806631} {\emph{DEMAND: Diverse Environments Multi-Channel Acoustic Noise Database}}.
\newblock The Journal of the Acoustical Society of America.

\bibitem[{{Mozilla Foundation}(2021)}]{commonvoice_es_v7_1}
{Mozilla Foundation}. 2021.
\newblock \href {https://commonvoice.mozilla.org/} {\emph{Common Voice Corpus (Spanish), version 7.1}}.
\newblock Mozilla Common Voice.

\bibitem[{{OpenSLR}(2016)}]{openslr_slr28_rirs_noises}
{OpenSLR}. 2016.
\newblock \href {https://www.openslr.org/28/} {\emph{Room Impulse Response and Noise Database (SLR28)}}.
\newblock OpenSLR.

\bibitem[{Solak(2019)}]{m_ailabs_speech_dataset}
Imdat Solak. 2019.
\newblock \href {https://www.caito.de/2019/01/03/the-m-ailabs-speech-dataset/} {\emph{The M-AILABS Speech Dataset}}.
\newblock M-AILABS.

\bibitem[{Veliche et~al.(2024)Veliche, Huang, Kochaniyan, Peng, Kalinli, and Seltzer}]{veliche2024towards}
Irina-Elena Veliche, Zhuangqun Huang, Vineeth~Ayyat Kochaniyan, Fuchun Peng, Ozlem Kalinli, and Michael~L Seltzer. 2024.
\newblock \href {https://arxiv.org/abs/2408.12734} {Towards measuring fairness in speech recognition: Fair-speech dataset}.
\newblock \emph{arXiv preprint arXiv:2408.12734}.

\end{thebibliography}

\section{Language Resource References}
\label{sec:lang_ref}
\bibliographystylelanguageresource{lrec2026-natbib}

\end{document}